\documentclass[pmlr,twocolumn,10pt]{jmlr} %

\usepackage{booktabs}
\usepackage{siunitx}

\usepackage{xcolor,colortbl}
\usepackage{color, colortbl} %
\usepackage{nicematrix}
\usepackage{enumitem}       %

\newcommand{\rev}[1]{\textcolor{black}{#1}}

\theorembodyfont{\upshape}
\theoremheaderfont{\scshape}
\theorempostheader{:}
\theoremsep{\newline}

\jmlrvolume{225}
\jmlryear{2023}
\jmlrsubmitted{LEAVE UNSET}
\jmlrpublished{LEAVE UNSET}
\jmlrworkshop{Machine Learning for Health (ML4H) 2023} %

\newcommand{\Sref}[1]{\S\ref{#1}}

\definecolor{lightBlue}{rgb}{0.78, 0.85, 1.0}%
\definecolor{lightOrange}{rgb}{0.88, 0.95, 1.0}%
\definecolor{lightRed}{rgb}{1.0, 0.85, 0.85}%
\usepackage{tcolorbox}%
\usepackage{multirow}

\newtcbox{\bluebox}{on line, box align=base, colback=lightBlue,colframe=white,size=fbox,arc=3pt, before upper=\strut, top=-2pt, bottom=-4pt, left=-2pt, right=-2pt, boxrule=0pt}

\newtcbox{\orangebox}{on line, box align=base, colback=lightOrange,colframe=white,size=fbox,arc=3pt, before upper=\strut, top=-2pt, bottom=-4pt, left=-2pt, right=-2pt, boxrule=0pt}

\newtcbox{\redbox}{on line, box align=base, colback=lightRed,colframe=white,size=fbox,arc=3pt, before upper=\strut, top=-2pt, bottom=-4pt, left=-2pt, right=-2pt, boxrule=0pt}

\newcommand{\dashifted}{\raisebox{0.5\depth}{\tiny$\downarrow$}}
\newcommand{\upshifted}{\raisebox{0.5\depth}{\tiny$\uparrow$}}

\newcommand{\dar}[1]{{\scriptsize\redbox{\dashifted{#1}}}}

\newcommand{\uab}[1]{{\scriptsize\bluebox{\upshifted{#1}}}}
\newcommand{\uao}[1]{{\scriptsize\orangebox{\upshifted{#1}}}}

\newcommand{\tJointTraining}{Joint}
\newcommand{\tStagedTraining}{Staged}

\newcommand{\mVotingFusion}{Best Voting}
\newcommand{\mEnsembleFusion}{Ensemble}
\newcommand{\mComplexFeatureFusion}{Feature}

 \title[Deep Multimodal Fusion for Surgical Feedback Classification]{Deep Multimodal Fusion for Surgical Feedback Classification}

\author{%
\Name{Rafal Kocielnik} \Email{rafalko@caltech.edu}\\
\addr California Institute of Technology, USA
\AND
\Name{Elyssa Y. Wong} \and \Name{Timothy N. Chu} 
\Email{eywong@usc.edu, tnchu@usc.edu}\\
\addr University of Southern California, USA
\AND
\Name{Lydia Lin} \Email{ljlin@usc.edu}\\
\addr University of Southern California \& California Institute of Technology, USA
\AND
\Name{De-An Huang} \Email{deahuang@nvidia.com}\\
\addr NVIDIA, USA
\AND
\Name{Jiayun Wang} \and \Name{Anima Anandkumar} \Email{peterw@caltech.edu, anima@caltech.edu}\\
\addr California Institute of Technology, USA
\AND
\Name{Andrew J. Hung} \Email{andrew.hung@cshs.org}\\
\addr Cedars-Sinai Medical Center, USA
}

\begin{document}

\maketitle

\begin{abstract}
Quantification of real-time informal feedback delivered by an experienced surgeon to a trainee during surgery is important for skill improvements in surgical training. Such feedback in the live operating room is inherently multimodal, consisting of verbal conversations (e.g., questions and answers) as well as non-verbal elements (e.g., through visual cues like pointing to anatomic elements).
In this work, we leverage a clinically-validated five-category classification of surgical feedback: \textit{``Anatomic''}, \textit{``Technical''}, \textit{``Procedural''},
\textit{``Praise''} and \textit{``Visual Aid''}. We then develop a multi-label machine learning model to classify these five categories of surgical feedback from inputs of text, audio, and video modalities.
The ultimate goal of our work is to help automate the annotation of real-time contextual surgical feedback at scale.
Our automated classification of surgical feedback achieves AUCs ranging from 71.5 to 77.6 with the fusion improving performance by 3.1\%. We also show that high-quality manual transcriptions of feedback audio from experts improve AUCs to between 76.5 and 96.2, which demonstrates a clear path toward future improvements.
Empirically, we find that the \emph{Staged} training strategy, with first pre-training each modality separately and then training them jointly, is more effective than training different modalities altogether. 
We also present intuitive findings on the importance of modalities for different feedback categories. This work offers an important first look at the feasibility of automated classification of real-world live surgical feedback based on text, audio, and video modalities.
\end{abstract}
\begin{keywords}
{Surgical feedback, Multimodality, Robot-Assisted Surgery, Deep Learning}
\end{keywords}

\begin{figure*}[t!]
  \begin{center}
\includegraphics[width=1.0\textwidth]{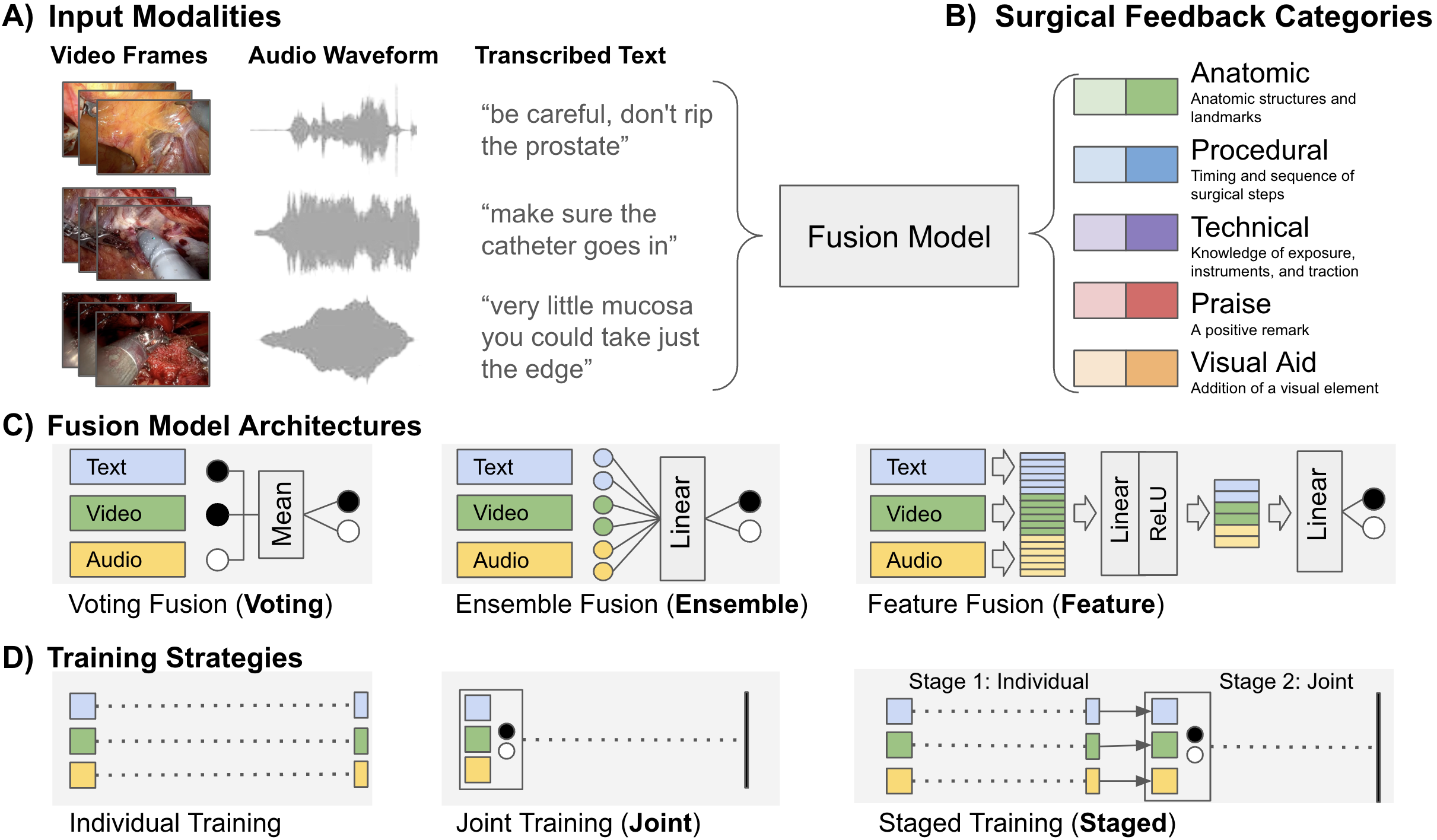}
  \caption{Overview of the work. Multimodal inputs consist of text, audio, and video (\textbf{A}) and 5 binary multi-label classification outputs adapted from a clinically validated framework introduced in \cite{wong2023feedback} (\textbf{B}). We explore model architectures (\textbf{C}) and training strategies (\textbf{D}) for improving the performance of surgical feedback classification using multimodal fusion.}
  \label{fig:fusion_main}
  \end{center}
  \vspace{-18.0pt}
\end{figure*}

\section{Introduction}
\noindent \textbf{Importance:} Real-time informal verbal feedback in surgical settings is pivotal not just for immediate correction and guidance but also for long-term proficiency and mastery \citep{agha2015role}. The quality of such feedback has been demonstrated to significantly influence intraoperative performance \citep{bonrath2015comprehensive}, profoundly impact surgical skill acquisition \citep{ma2022tailored} as well as trainee's sense of autonomy \citep{haglund2021surgical}. It also has broader implications for the overall surgical training paradigm. Despite the inherent challenges posed by the unstructured and personalized nature of surgical feedback, it's undeniable that a systematic approach to understanding it is the linchpin to refining and enhancing surgical training.

\vspace{2.0pt}
\noindent \textbf{Challenges:} However, quantifying and conducting a systematic analysis of the properties of real-world surgical feedback presents notable challenges. We, therefore, adopt a recent clinically validated classification system for surgical feedback that has been shown to offer high reliability and generalizability as well as practical utility \citep{wong2023feedback}. However, their system requires manual annotations of surgical feedback, which is time and resource-demanding. 
This is primarily due to the necessity for expertise in comprehending both the surgical context and the feedback's intent \citep{agha2015role}. Furthermore, feedback delivery in the live operating room is inherently multimodal and adds to the complexity. The delivery encompasses verbal conversations, non-verbal appraisals, and visual cues. 

\vspace{2.0pt}
\noindent\textbf{Approach:} We explore automated intraoperative surgical feedback classification with machine learning techniques in this pilot study. %
Specifically, we leverage multi-modal inputs composed of text, audio, and video (Fig. \ref{fig:fusion_main}-A) in order to perform binary multi-label classification of surgical feedback into 5 components (Fig. \ref{fig:fusion_main}-B). In our experiments we systematically vary 2 dimensions: 1) complexity of the fusion model architecture (Fig. \ref{fig:fusion_main}-C) and 2) training strategy (Fig. \ref{fig:fusion_main}-D). We arrive at an optimal \emph{Staged Fusion} approach which starts with independent training of each modality and continues with training modalities jointly. This approach helps mitigate the dominance of one modality that can suppress extracting information from other modalities.

\vspace{2.0pt}
\noindent \textbf{Findings:} We summarize our findings as follows:
\begin{itemize}[leftmargin=*, itemsep=0.0mm, topsep=5pt]
    \item We achieve Areas under the ROC Curve (AUCs) varying from 71.5 to 77.6 with automated surgical feedback classification (Table \ref{tab:res_table}). 
    
    \item We further show that manual transcription of specialized surgical feedback by experts, though costly, further improves AUCs to between 76.5 and 96.2, indicating a path to further improvements.
    
    \item We find that the training process is more important for fusion effectiveness (3.1\% gain) than model architecture (1.1\%) in ablation studies.
    \item We confirm our intuition that video modality is most important for the classification of \textit{``Visual Aid''} feedback, while emotion extracted from audio is important for the detection of \textit{``Praise''}.
  
\end{itemize}

\begin{table*}[htb]
\centering
\label{tab:feedback}
\small{
\resizebox{\textwidth}{!}{
\begin{tabular}{@{} l   @{\hskip 0.03in} | @{\hskip 0.05in} l}
\toprule
 \textbf{Feedback} & \textbf{Description} \\

 \midrule

 Anatomic & Familiarity with anatomic structures and landmarks. \\
 
 Procedural & Pertains to timing and sequence of surgical steps.  \\

 Technical & Performance of discreet task with appropriate knowledge of exposure, instruments, traction, etc.  \\

 Praise & A complementary remark  \\

 Visual Aid & Addition of visual element to direct trainee's attention or focus \\

\bottomrule

\end{tabular}

}
}
\caption{Categories of surgical feedback adapted from recent clinically validated classification system introduced in \cite{wong2023feedback}}.
\vspace{-12pt}
\end{table*}

\noindent \textbf{Contributions:} Our main contributions include:%
\begin{itemize}[leftmargin=*, itemsep=0.0mm, topsep=5pt]

    \item To the best of our knowledge, we are the first to explore the feasibility of the automated classification of live surgical feedback.

    \item We systematically explore model architectures and training strategies for multi-modal fusion in a novel context of real-world live surgical feedback. The emphasis on training strategy distinguishes our approach as significantly novel, given that most prior work focused on exploring model architectures.

\end{itemize}

\section{Background and Related Work}

\paragraph{Feedback in Robot-Assisted Surgery}. 
\cite{wong2023feedback} first report on the development of a manual classification system for verbal feedback during robot-assisted surgery. \rev{This work also demonstrates the reliability, generalizability, and utility of this manual classification system. It specifically shows that using the proposed feedback categorization it is possible to detect significant differences in feedback type frequency and subsequent trainee reactions based on surgeon experience level and the surgical task being performed. For example, technical feedback with a visual component was associated with an increased rate of trainee behavioral change or verbal acknowledgment responses. Hence we adopt this classification system as it offers a tangible link between feedback and subsequent trainee behavior.}

To the best of our knowledge, there exists no prior work on automated surgical feedback classification. Our work pioneers predicting real-time verbal feedback for robotic-assisted surgery with multi-modal sensory inputs.

\paragraph{Deep Learning for Multi-Modality Data}. Prior work mostly focused on fusing visual modalities but not the importance of training strategies. \cite{boulahia2021early} explore \textit{early}, \textit{intermediate}, and \textit{late} fusion for general activity recognition. Their method focuses on visual channels and is not directly applicable to surgical feedback which includes text and audio modalities. We borrow the late fusion concept from their work, but expand on aspects of model complexity and training strategy. \cite{li2020oscar} align text and image modalities for image captioning task. This fusion approach aims to generate output in one modality based on input from other modalities, which is substantially different than our task. \cite{walsman2019early} focus on the fusion of visual channels for the scene and goal representation in robotic vision. Their work applies fusion in 3D simulated setting with clear and distinct objects, which are not present in our context. In the medical domain, \cite{narazani2022pet} explore fusion for PET and MRI visual modalities. Their work, again focuses on visual channels only and reports no gains from the proposed fusion approaches. In contrast, our research systematically investigates a range of multi-modal fusion techniques and training strategies.

\section{Methods}
\subsection{Data Acquisition}

\begin{figure*}[t!]
  \begin{center}
\includegraphics[width=1.0\textwidth]{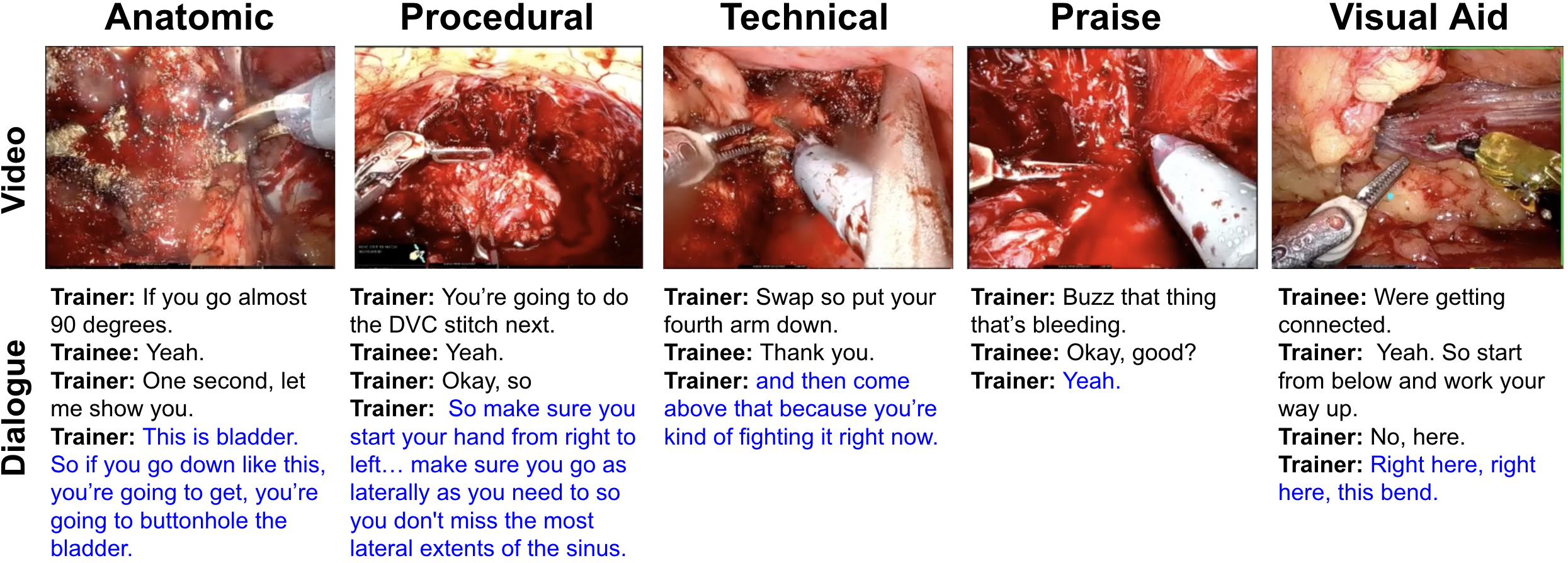}
\vspace{-18.0pt}
  \caption{Examples of video (frame) along with the dialogue between trainee (feedback recipient) and attending surgeon (feedback provider) from different surgical cases in our dataset.}
  \label{fig:feedback_showcase}
  \end{center}
  \vspace{-18.0pt}
\end{figure*}

\begin{figure*}[t!]
  \begin{center}
\includegraphics[width=1.0\textwidth]{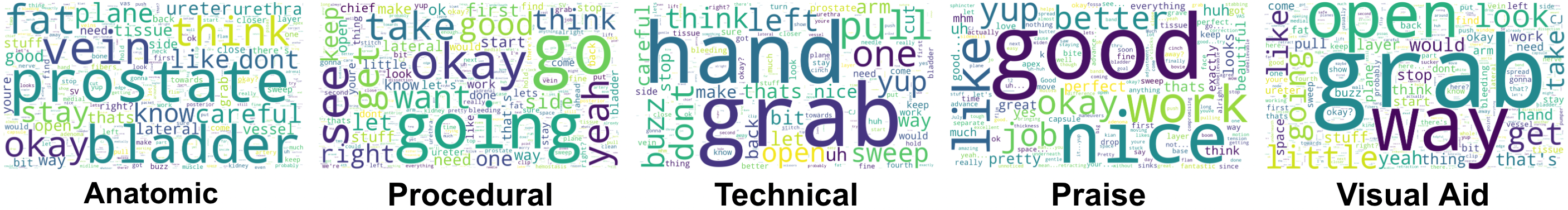}
\vspace{-18.0pt}
  \caption{Most frequent words used in the delivery of each type of feedback visualized via word clouds. The larger the word, the more frequently it has been used in this category of feedback.
For example, \emph{Anatomic} feedback includes words related to physical structures like ``prostate'' and ``bladder'', while \emph{Technical} feedback  includes words describing the use of instruments like ``grab'' and ``hand''.}
  \label{fig:feedback_word_clouds}
  \end{center}
  \vspace{-20.0pt}
\end{figure*}

\begin{table}[t!]
\centering
\small
\begin{tabular}{lrrr}
\toprule
 Component & Count & Count/Case & Word count  \\
 \midrule

 Anatomic & 1104 & $35.6\pm23.0$ & $11.0\pm7.7$ \\
 Procedural & 817 & $26.4\pm14.9$ & $9.8\pm8.0$ \\
 Technical & 3223 & $104.0\pm67.7$ & $8.1\pm6.8$ \\
 Praise & 262 & $9.0\pm\;\:8.4$ & $3.6\pm3.7$ \\
 Visual Aid & 303 & $11.7\pm10.2$ & $10.5\pm6.7$ \\
 \midrule
 \textbf{Any feedback} & 3912 & $126.2\pm72.0$ & $8.1\pm6.8$ \\[0pt]
 
 \bottomrule

\end{tabular}
\caption{Statistics per surgical feedback category including total instances, instances per individual surgical case as well as mean word count of transcribed feedback text. \textbf{Any feedback} refers to feedback of any type. One feedback might have multiple labels.}
\label{tab:dataset-stats}
\end{table}

We used a dataset of real-life feedback delivered by trainers to trainees during live robot-assisted surgery cases from \cite{wong2023feedback}. Trainers were defined as those providing feedback and trainees were those receiving feedback while actively operating on the surgeon console. This feedback has been recorded using wireless microphones worn by the surgeons and video capturing the surgeon's point-of-view (i.e., endoscope camera view). Video and audio were recorded synchronously with an external recorder. All surgeries were performed using da Vinci Xi surgical robotic system \citep{dimaio2011vinci}. The feedback instances were timestamped and manually transcribed from audio recordings. Feedback instance has been defined as trainers' utterances meant to alter or approve trainee behavior. The dataset contains 3912 individual instances of feedback as shown in Table \ref{tab:dataset-stats}.

\subsection{Surgical Feedback Categorization}

Two medical students were involved in feedback identification and transcription. Manual transcription included only utterances from the attending surgeon providing feedback. Any utterances by trainees or unrelated conversations were not transcribed. 

This feedback has been categorized using surgical feedback quantification framework introduced by \cite{wong2023feedback}. This categorization scheme has been shown to offer high reliability and generalizability as well as practical utility in the clinical setting. The five feedback dimensions from this framework along with their definitions are presented in Table \ref{tab:feedback}. The categories are non-exclusive. Further details of the annotation can be found in \cite{wong2023feedback}.

Examples of aligned video frames and audio transcriptions are shown in Fig. \ref{fig:feedback_showcase}. Dialogue is very important in feedback categorization, whereas video offers supplementary sources, but similar feedback can be delivered in different visual contexts. Fig. \ref{fig:feedback_word_clouds} shows the most frequent words for each feedback category as word clouds where the larger the word, the more frequently it appears in the underlying feedback instances. \emph{Anatomic} type of feedback most frequently includes words related to physical structures such as ``prostate'', ``bladder'', and ``vein''. At the same time, \emph{Technical} feedback frequently includes words such as ``grab'' and ``hand'', ``pull' referring to the use of instruments.

\subsection{Speaker Diarization and Automated Speech Recognition}
\label{sec:asr_diarization}
In addition to manual transcription, we performed \emph{Automated Speech Recognition (ASR)} using pre-trained Whisper medium model introduced in \cite{radford2022whisper}. This model was pre-trained on 680k hours of labeled English-only speech data specifically for speech recognition. Speech data was annotated using large-scale weak supervision. Given, the interactive dialogue-like structure of the exchanges around and leading to feedback (see Fig. \ref{fig:feedback_showcase}), we further applied speaker diarization, the concept of partitioning speech from different speakers in a single audio clip, using \emph{Pyannote} \citep{Bredin2021, Bredin2020}. This was done to provide more context about feedback such as the speaker and the conversations before and after the feedback delivery. Speaker diarization was paired with the ASR to transcribe each separate segment of audio. 

\subsection{Individual-Modality-Input Models}
We leverage pre-trained transformer models to extract information from each individual modality. 

\noindent \textbf{Text:} We fine-tune \emph{BERT} base model with 110M parameters introduced by \cite{devlin2018bert}. The model has been pre-trained on general-knowledge text including BooksCorpus and English Wikipedia. We also experiment with specialized text models pre-trained on biomedical datasets including \emph{BioBert} \citep{lee2020biobert} and \emph{BioClinicalBert} \citep{alsentzer2019publicly}. However, no noticeable improvement in performance has been observed, which we attribute to the relatively casual and conversational nature of the feedback with only occasional use of specialized vocabulary.

\noindent \textbf{Audio:} We fine-tune \emph{Wave2Vec} base model with 95M parameters introduced by \cite{baevski2020wav2vec}. We specifically use a model pre-trained on emotion recognition tasks from \textit{``SUPERB''} dataset \citep{yang2021superb}. This model extracts features related to the emotion in the delivery of feedback from audio and is different than text transcription.

\noindent \textbf{Video:} We fine-tune \emph{VideoMAE} base model with 86M trainable parameters introduced by \cite{tong2022videomae}. This model is an extension of Masked Auto Encoders introduced by \cite{he2022masked} from images to video. We use a model pre-trained on Kinetics-400 dataset \citep{kay2017kinetics} containing video clips of 400 human action classes.

\subsection{Model Architectures of Multi-Modality Fusion}
We explore different variants of late fusion (Fig. \ref{fig:fusion_main}-C) varying the model complexity from a simple majority vote to feature fusion with additional layers.

\definecolor{highlight}{rgb}{1.0,0.90,0.8}	
\definecolor{Gray}{gray}{0.96}

\begin{table*}[t]
\centering
\small{
\begin{NiceTabular}{@{} l | l @{\hskip 0.20in} l @{\hskip 0.20in} l @{\hskip 0.20in} l @{\hskip 0.20in} l @{\hskip 0.15in} |c  @{\hskip 0.07in} }[colortbl-like]
\CodeBefore 
\Body
 \textbf{Model} & \textbf{Anatomic} & \textbf{Procedural} & \textbf{Technical} & \textbf{Praise} & \textbf{Vis. Aid} & \textbf{Mean \%}\\

 \midrule

\rowcolor{Gray}
Text (Manual)$^1$ & $81.5_{3.3}$ & $69.3_{3.6}$ & $74.3_{1.9}$ & $95.2_{2.4}$ & $78.4_{3.1}$ \\

Text (ASR)$^2$ & $70.3_{3.2}$ & $65.7_{4.7}$ & $66.5_{4.0}$ & $76.2_{8.5}$$^{\dagger}$ & $66.7_{6.8}$ \\
Audio (Emotion) & $67.3_{0.3}$ & $61.8_{2.3}$ & $67.2_{2.8}$ & $67.3_{6.2}$$^{\dagger}$ & $61.2_{5.5}$\\
Video & $65.7_{2.1}$ & $64.0_{2.8}$ & $66.0_{0.5}$ & $57.0_{2.2}$ & $73.0_{6.4}$$^{\ddagger}$\\
\midrule

\multicolumn{7}{c}{\rowcolor{Gray}\textbf{$^1$Fusion Using Manual Transcription}} \\
\midrule

\rowcolor{Gray}
\mVotingFusion{}  & $79.7_{2.0}$ \dar{2.2\%} & $72.0_{2.2}$ \uab{3.8\%} & $74.2_{5.0}$ \dar{0.2\%} & $76.9_{4.3}$ \dar{19.3\%} & $78.4_{1.3}$ \dar{0.0\%} & \dar{3.6\%} \\[5pt]

\rowcolor{Gray}
\tJointTraining{}-\mEnsembleFusion{} & $81.7_{3.3}$ \uao{0.2\%} & $72.3_{0.8}^{*}$ \uab{4.3\%} & $74.7_{4.4}$ \uao{0.4\%} & $95.5_{1.1}$ \uao{0.3\%} & $82.2_{1.7}^{*}$ \uab{4.9\%} & \uab{2.0\%}\\

\rowcolor{Gray}
\tStagedTraining{}-\mEnsembleFusion{} & \underline{$86.0_{2.6}^{*}$} \uab{5.5\%} & \underline{$76.5_{2.3}^{*}$} \uab{10.3\%} & $78.8_{3.8}^{*}$ \uab{6.1\%} & \underline{$96.2_{1.9}^{*}$} \uao{1.0\%} & \underline{$86.1_{1.4}^{*}$} \uab{9.8\%} & \uab{6.5\%}\\[5pt]

\rowcolor{Gray}
\tJointTraining{}-\mComplexFeatureFusion{} & $81.8_{1.5}$ \uao{0.4\%} & $72.2_{5.6}$ \uab{4.1\%} & $76.2_{0.8}^{*}$ \uab{2.5\%} & $95.5_{1.5}$ \uao{0.3\%} & $80.6_{2.5}$ \uab{2.8\%} & \uab{2.0\%}\\

\rowcolor{Gray}
\tStagedTraining{}-\mComplexFeatureFusion{} & \underline{$86.0_{1.8}^{*}$} \uab{5.5\%} & $76.3_{2.8}^{*}$ \uab{10.1\%} & \underline{$80.3_{4.9}^{*}$} \uab{8.1\%} & $95.9_{1.0}$ \uao{0.7\%} & $85.8_{1.7}^{*}$ \uab{9.4\%} & \uab{6.8\%}\\

\midrule

\multicolumn{7}{c}{\textbf{$^2$Fusion Using Automated Transcription (ASR) and Speaker Diarization}} \\
\midrule

\mVotingFusion{}  & $69.2_{0.3}$ \dar{1.7\%} & $63.8_{1.9}$ \dar{2.8\%} & $68.5_{2.7}$ \uab{1.2\%} & $70.5_{3.4}$ \dar{7.5\%} & $70.5_{3.6}$ \dar{3.4\%} & \dar{2.8\%} \\[5pt]

\tJointTraining{}-\mEnsembleFusion{} & $70.5_{0.9}$ \uao{0.2\%} & $65.8_{1.3}$ \uao{0.3\%} & $68.5_{1.8}$ \uab{1.4\%} & $75.2_{1.8}$ \dar{1.2\%} & $76.5_{3.9}^{*}$ \uab{4.9\%} & \uab{1.1\%}\\

\tStagedTraining{}-\mEnsembleFusion{} & \underline{$71.7_{3.3}^{*}$} \uab{1.9\%} & \underline{$71.5_{1.7}^{*}$} \uab{8.9\%} & $69.2_{5.4}$ \uab{2.2\%} & \underline{$76.8_{8.2}$} \uao{0.9\%} & $74.0_{3.7}$ \uab{1.5\%} & \uab{3.1\%}\\[5pt]

\tJointTraining{}-\mComplexFeatureFusion{} & $68.3_{2.8}$ \dar{2.8\%} & $66.3_{1.5}$ \uao{1.0\%} & $66.5_{1.0}$ \dar{1.7\%} & $75.6_{2.7}$ \dar{0.8\%} & $76.0_{8.5}$ \uab{4.1\%} & \uao{0.0\%}\\

\tStagedTraining{}-\mComplexFeatureFusion{} & $70.5_{2.5}$ \uao{0.2\%} & $66.7_{3.0}$ \uab{1.5\%} & \underline{$72.2_{2.6}^{*}$} \uab{6.7\%} & $76.2_{7.4}$ \uao{0.0\%} & \underline{$77.6_{5.8}^{*}$} \uab{6.4\%} & \uab{3.0\%}\\

\bottomrule

\end{NiceTabular}
}
\caption{Feedback classification results based on Manual Transcription - \emph{Text (Manual)} and Automated Speech Recognition - \emph{Text (ASR)}. \textbf{Mean \%} refers to the average gain of the model taking multi-modality over the best performing single modality input. The subscripts are the standard deviation of different runs. \textbf{$^{*}$} indicates a statistically significant gain compared to the best individual modality model at p$<$0.05. Note that for \textit{\textbf{Praise}}, due to the information contained in particular modalities, is expected that $^{\dagger}$\emph{Text} input only leads to high classification performance while video only leads to relatively low performance. Similarly for \textit{\textbf{Visual Aid}} due to reliance on visual pointing, the $^{\ddagger}$\emph{Video} modality is expected to perform particularly well.
See Fig. \ref{fig:fusion_examples} for details.
}

\label{tab:res_table}
\vspace{-12.0pt}
\end{table*}

\noindent\textbf{Voting Fusion (\mVotingFusion{}):} In this architecture, each modality model predicts the label independently (e.g., whether feedback component is \emph{``Anatomic''} or \emph{``Non-anatomic''} based on video only). The prediction given by the majority of models (i.e., at least 2 out of 3 models), is used as the final label for the fusion model. We further explore voting fusion via max of model predictions (i.e., at least 1 model predicts a positive label). We report the best of these voting approaches in our results.

\noindent \textbf{Ensemble Fusion (\mEnsembleFusion{}):} In this architecture, each model returns a size 2 vector representation of the modality. These reduced representations are combined via a linear 6x2 layer which weights each modality and returns the probability of each class (e.g., \emph{``Anatomic''} or \emph{``Non-Anatomic''}) as the final fusion output. Compared to \mVotingFusion{} approach, the \mEnsembleFusion{} architecture can learn the optimal weighting for combining the representations from each individual modality.

\noindent \textbf{Feature Fusion (\mComplexFeatureFusion{}):} In this architecture, we extract much richer representations from each modality in the form of 256-dimension vector. This can help capture more detailed information, but may also add complexity to the learning process. The representations are concatenated into one 756-dim vector and passed via 2 fully-connected linear layers that reduce the dimensions to 96 and finally 2 in a funnel fashion. This sequential architecture is augmented with ReLu activation and additional dropout in between. The additional steps can help the model calculate intermediate fusion features.

\subsection{Training Strategies of Multi-Modality Fusion}
We explore 3 training strategies as depicted in Fig. \ref{fig:fusion_main}-D: 1) Individual training of each modality, 2) Joint training (J) of all modalities, 3) Staged training (S), which starts with individual training followed by further joint training.

\noindent\textbf{Individual Training:} Each modality model is trained independently for the same number of epochs. Each modality also makes an independent prediction about the final label. This setup offers a simple no-fusion baseline. We further use the independently trained models with the voting fusion model (\mVotingFusion{}) to offer the basic fusion baseline.

\noindent\textbf{Joint Training (\tJointTraining{}):} The individual modality models are combined under one fusion architecture (\mEnsembleFusion{} or \mComplexFeatureFusion{}) and trained jointly for the whole duration of the training. This approach allows the fusion model to learn how to extract relevant information from each modality simultaneously and possibly also learn the differences between modalities relevant to the task.

\noindent\textbf{Staged Training (\tStagedTraining{}):} The models for each modality are first pre-trained independently on the same task for half of the training time (Stage 1 \textit{``Individiual''}). Then the pre-trained models are combined under the fusion model (\emph{\mEnsembleFusion{}} or \emph{\mComplexFeatureFusion{}}) and trained further jointly for the remainder of the training time (Stage 2 \textit{``Joint''}). The first stage helps each model extract relevant information from its modality without interference from other modalities. Extraction of such information from less predictive modalities can otherwise be suppressed. 

\subsection{Evaluation Schemes and Setups}
\label{sec:methods-eval}
We obtain baselines for each individual modality by fine-tuning models for the same number of epochs and reporting the top AUC score obtained on the test set. 
For all our experiments, we use label-balanced datasets for each feedback dimension obtained via random downsampling of the majority class. We use an 80\%/20\% random train/test split and perform each experiment 3 times with a controlled random seed and report mean AUC as well as standard deviation. Dimension-specific label balancing leads to variable dataset size for each dimension, specifically: \emph{Anatomic} ($N$=2208), \emph{Procedural}($N$=1634), \emph{Technical}($N$=1378), \emph{Praise} ($N$=524), \emph{Visual Aid} ($N$=606). 

\rev{We further compare the performance of the fusion approaches to the best-performing individual modality model using McNemar's non-parametric statistical test as suggested in \cite{dietterich1998approximate} and further adapted to the settings involving expensive deep learning setups by \cite{vanwinckelen2012estimating}. We use a Python implementation of McNemar's test provided in \cite{raschkas_2018_mlxtend}.
}

\subsection{Data Processing and Model Training}
We trim a 10-second video with audio information when human-annotated feedback appears. This includes 5 seconds before (to capture context) and 5 seconds after (to capture delivery) the feedback onset. We preprocess the video by downsampling the resolution to $320\times250$ and extracting 16 randomly uniformly sampled frames. We preprocess the audio by downsampling it to 16kHz mono. 
We train all the fusion models for a total of 20 epochs with the same initial learning rate (LR) of $5\text{e}-6$, Adam optimizer, and a scheduler that reduces LR when an AUC has stopped improving for 2 epochs (patience) with a reduction factor of 0.5. We use a batch size of 2 with a gradient accumulation of 10.

\section{Results}

\subsection{Feedback Classification Results}
Table \ref{tab:res_table} summarizes the results of the classification of feedback components. The top rows report AUCs for individual modalities. We include 2 versions of transcribed text from audio. \emph{Text(Manual)} - costly manual transcription by human experts, \emph{Text(ASR)} - automated transcription from audio using ASR and Speaker Diarization as described in \Sref{sec:asr_diarization}. Text modality itself is highly predictive for each component. In the subsequent two sections of the table we report AUCs for multi-modal fusion approaches relying on high quality, but costly manual transcriptions - \emph{``Fusion Using Manual Transcription} and same fusion approaches relying on automated transcription from audio - \emph{``Fusion Using Automated Transcriptions (ASR) and Speaker Diarization''}.

In each row we report the AUCs for different fusion approaches. The \mVotingFusion{} is a majority vote fusion baseline. The following rows report results for joint (\tJointTraining{}) and staged (\tStagedTraining{}) training approaches of the same architecture \emph{Ensemble Fusion} (\mEnsembleFusion{}) model. The last two rows report the results of joint (\tJointTraining{}) and staged (\tStagedTraining{}) training for \emph{Feature Fusion} (\mComplexFeatureFusion{}) model. Next to each AUC score for fusion approaches, we report relative gain or loss with respect to the highest AUC from individual modalities. We also underscore the highest AUC per each feedback component across all models.

\paragraph{Varying, but Consistent Gains from Fusion:} The top AUC for automated classification of \textit{``Praise''} is high at 76.8, and fusion provides the least gain of 0.9\% for this component. This is likely because praise can be delivered in different visual contexts and hence video does not provide much more information. The AUC is also high, at 77.6, for \textit{``Visual Aid''}. In this case, the gain from incorporating video is substantial at 6.4\%. This is due to the visually observable pointing associated with this feedback component. For \textit{``Anatomic''}, \textit{``Procedural''}, and \textit{``Technical''} the top AUCs are 71.7, 71.5, and 72.2 respectively. The best fusion approach provides a noticeable gain for these components of between 1.9\% and 8.9\%. 

\paragraph{Staged Training Outperforms Other Approaches:} Looking at the results of staged training (\emph{\tStagedTraining{}-\mEnsembleFusion{}} and \emph{\tStagedTraining{}-\mComplexFeatureFusion{}}) in comparison to other fusion approaches, the staging outperforms them all. We observe a mean gain of 3.0\% to 6.8\% across fusion relying on automated and manual transcription respectively. This is compared to smaller gains of just 2.0\% or even no gain for the joint training with the exact same model architectures (\emph{\tJointTraining{}-\mEnsembleFusion{}} and \emph{\tJointTraining{}-\mComplexFeatureFusion{}}). At the same time, the simple majority vote fusion (\emph{\mVotingFusion{}}) leads to AUC loss in 4 out of 5 dimensions over the best individual modality. 

\rev{Given relatively high standard deviations we examine the statistical significance of the observed gains as described in \Sref{sec:methods-eval}. Statistically significant gains are marked with ``$^{*}$'' in Table \ref{tab:res_table}. For fusion relying on manual transcriptions, the best-performing fusion architecture offers statistically significant gains over the best-performing single modality model for all feedback components: \textit{Anatomic} (M$=$4.5, p$<$0.01), \textit{Procedural} (M$=$7.2, p$<$0.01), \textit{Technical} (M$=$6.0, p$<$0.01), \textit{Praise} (M$=$1.0, p$<$0.05), and \textit{Visual Aid} (M$=$7.7, p$<$0.01); where M denotes the best mean absolute AUC gain.}

\rev{For fusion relying on automated transcription, we observe statistically significant gains from the best-performing architecture for 4  out of 5 feedback components:
\textit{Anatomic} (M$=$1.4, p$<$0.05),
\textit{Procedural} (M$=$5.7, p$<$0.01),
\textit{Technical} (M$=$5.0, p$<$0.01),
\textit{Praise} (M$=$0.6, p$=$0.58, n.s.), \textit{Visual Aid} (M$=$4.6, p$<$0.05).
}

\begin{figure*}[t!]
  \begin{center}
\includegraphics[width=1.0\textwidth]
{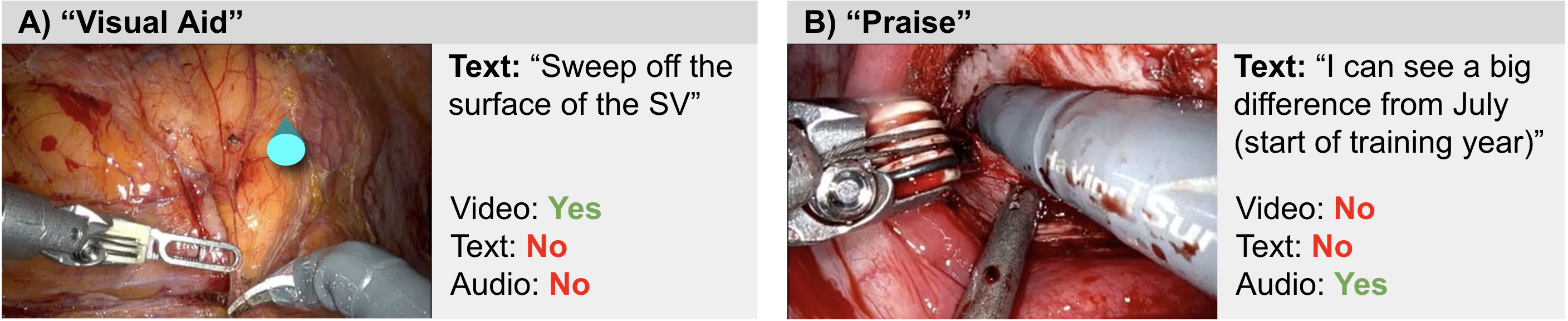}
  \caption{Examples of video clips where a single modality does not contain enough information to make a correct classification. \textbf{A}) The verbal feedback itself is not enough to determine whether a trainer is pointing to anything specific. It is necessary to look at video modality and the use of a pointer (teal cone) to make a correct determination. Please note that we enlarged the pointer for visual clarity. \textbf{B}) The text itself is too ambiguous to determine whether the feedback is positive or not. Additional information from the tone of voice provides the necessary distinction.}
  \label{fig:fusion_examples}
  \end{center}
  \vspace{-16.0pt}
\end{figure*}

\paragraph{Intuitive Value of Individual Modalities:} We further note intuitive patterns in the predictive value of individual modalities per feedback components. For \emph{``Praise''}, text and audio modalities alone achieve relatively higher AUCs of 76.2 and 67.3 respectively compared to video with AUC of 57.0. This is intuitive as this component captures the feedback delivery and can contain emotional undertones, but praise can be delivered in any visual context. On the other hand, for \emph{``Visual Aid''} video alone achieves a high AUC of 73.0, while audio alone achieves a much lower AUC of 61.2. This is again intuitive as this component captures the surgeon using a visual aid in the form of a cursor or surgical instrument as a pointer (see Fig. \ref{fig:fusion_examples}-A).

\paragraph{Manual v.s. Auto Transcription:} 
Table \ref{tab:res_table} shows that the experiments leveraging fusion using manual expert-provided transcriptions offer higher AUCs ranging from 76.5 to 96.2. This represents a 15.3\% average improvement over fusion relying on fully automated ASR transcription and speaker diarization. These results offer a likely upper bound for classification performance as well as show the potential of improving the processing of audio. However, obtaining manual expert transcriptions of live surgical feedback is costly. While there exist services for medical text transcription \citep{Princeto52:online}, they are still costly and may not offer the same quality for specialized surgical domains.

\subsection{Additional Analysis}
To better understand the value of fusion, we manually inspect several examples of disagreements between simple majority vote fusion, the true label, and the prediction from the best-performing fusion model. In Fig. \ref{fig:fusion_examples} we show two illustrative examples of such disagreements. In A) we show an example of \textit{``Visual Aid''} component classification, where neither text nor audio provides the correct label. Looking at the video, it is clear that the trainer is using a visual pointer, but pointing can also happen using instruments. In example B) for classification of \textit{``Praise''} the feedback text by itself is insufficient to correctly determine if the feedback is intended to be positive. The inclusion of delivery aspects from audio is important in that case. 

\rev{At scale, we quantified the impact on Precision and Recall. In the case of fusion relying on manual transcription, for \textit{``Visual Aid''} the best fusion model improved Precision by 15.5\% and Recall by 0.6\% compared to the Best Voting (see Appendix \ref{apd:additonal-metrics}). It also improved Precision by 10.5\% and Recall by 9.2\% compared to the best single modality. A similar impact can be observed for \textit{``Praise''} with improvement in Precision by 21.6\% and Recall by 35.1\% compared to baseline Best Voting and improvements in Precision by 1.9\% and Recall by 0.7\% compared to the best single modality. Similar trends can be seen in fusion relying on automated transcription. For \textit{``Visual Aid''} there is an improvement in Precision by 10.0\% and in Recall by 13.2\% when using the best fusion approach compared to Best Voting. Similarly, compared to the best single modality, fusion improves Precision by 6.7\% and Recall by 8.1\%. Enhanced Precision indicates fewer false positives, affirming the relevance of the identified feedback instances of a particular type. Improved Recall signifies fewer missed authentic feedback cases of that type. Improvement in both Precision and Recall underscores not only an increase in the accurate detection of specific feedback types but also a broader and more reliable capture of feedback components.}

\vspace{-4pt}
\section{Discussion}
We thoroughly evaluate different multi-modality fusion architectures and training strategies across 3 data splits and 5 surgical feedback classification dimensions. 
The low effectiveness of a simple majority vote gives us insights into the manner in which the modalities need to be combined. It seems important to have a number of trainable parameters in order to learn how to combine the information across modalities. We gain evidence for this via examination of disagreements between \emph{\mVotingFusion{}} and the staged fusion setting, which shows that improvements are based on \rev{both precision and recall scores.}

Further increase in the number of trainable parameters does not translate to improvements. It is likely that only a limited complexity is needed to relate the modalities effectively. We did not freeze any of the individual model weights and these models are themselves complex. 

We introduce staged training to address the issue of dominance of the text modality over other modalities. %
We observe that this approach led to the highest gain across classifications of all the feedback components irrespective of the fusion model complexity. This shows the importance of considering the training process itself for fusion, while most of the prior work focused on model architectures.

\rev{We further note that the automated multimodal classification we introduced in this work is based on a clinically validated manual system from \cite{wong2023feedback}. As such, these classification dimensions have been shown to be generalizable across 6 types of surgical procedures. They have also been shown to predict significant differences in surgeon experience level, the surgical task being performed, as well as the likelihood of behavioral adjustment observed among trainees (a measure of feedback effectiveness). This further shows the practical real-world utility of the automation of this classification system through the novel deep multimodal fusion approach we proposed.}

We note several future directions. 
First, the quantification of surgical feedback is an important first step towards generating the optimal feedback automatically using retrieval or generative models in the future \citep{laca2022using}.
Second, in this study, we experimented with both manual and automated transcription of feedback from audio. We show that manual transcription, which requires substantial effort, offers better performance. Further experiments should try to improve the performance of automated transcription \citep{moore2015automated}. Finally, our individual models are all transformer architectures capable of unsupervised pre-training, which could improve the overall performance even further.

\vspace{-10pt}
\section{Conclusion}
We present the first work to explore an automated classification of components of real-world informal live surgical feedback using a clinically validated classification scheme. We show that it is feasible to classify components of such feedback with promising AUCs varying from 71.5 up to 77.6. Secondly, we show that this feedback is indeed inherently multi-modal and fusion can meaningfully improve AUC by as much as 8.9\%. Third, we show that the multi-modal fusion through staged training is more effective than the fusion model architecture itself. This work provides important insights into the importance of training strategy for effective multi-modal fusion. We open up opportunities for quantification of surgical feedback at scale from text, audio, and video recordings, which can lead to improvements in surgical training and outcomes.

\acks{
\rev{Research reported in this publication was supported by the National Cancer Institute of the National Institutes of Health under Award Number R01CA251579 and R01CA273031. The content is solely the responsibility of the authors and does not necessarily represent the official views of the National Institutes of Health.}
}

\bibliography{kocielnik23}

\appendix

\section{Additional Classification Metrics}\label{apd:additonal-metrics}

In Table \ref{tab:add_res_vis_aid} we present additional F1-binary, Precision, and Recall metrics from \textit{Visual Aid} feedback component classification. Tables \ref{tab:add_res_praise}, \ref{tab:add_res_anatomic}, \ref{tab:add_res_procedural}, and \ref{tab:add_res_technical} contain additional metrics for \textit{Praise}, \textit{Anatomic}, \textit{Procedural}, and \textit{Technical} components respectively.

\begin{table}[h!]
\centering
\small
\begin{tabular}{lrrr}

\toprule
 Model & F1-binary & Precision & Recall  \\
 \midrule

 Text (Manual) & 78.05 & 78.94 & 77.59 \\
 Text (ASR) & 65.50 & 67.89 & 64.48 \\
 Audio (Emotion) & 60.83 & 62.21 & 60.11 \\
 Video & 73.19 & 72.64 & 73.77 \\
 \midrule
    \multicolumn{4}{c}{\textbf{Fusion using Manual Transcription}} \\
 \midrule

 \mVotingFusion{} & 79.58 & 76.56 & 84.16 \\[5pt]
 
 \tJointTraining{}-\mEnsembleFusion{} & 81.90 & 83.55 & 80.33 \\
 \tStagedTraining{}-\mEnsembleFusion{} & 85.84 & 87.27 & 84.70 \\[5pt]
 
 \tJointTraining{}-\mComplexFeatureFusion{} & 79.81 & 83.87 & 76.50 \\
 \tStagedTraining{}-\mComplexFeatureFusion{} & 86.10 & 84.45 & 97.98 \\

 \midrule
    \multicolumn{4}{c}{\textbf{Fusion using ASR Transcription}} \\
 \midrule

 \mVotingFusion{} & 70.47 & 70.48 & 70.49 \\[5pt]
 
 \tJointTraining{}-\mEnsembleFusion{} & 76.42 & 76.93 & 75.96 \\
 \tStagedTraining{}-\mEnsembleFusion{} & 72.86 & 77.34 & 69.40 \\[5pt]
 
 \tJointTraining{}-\mComplexFeatureFusion{} & 73.20 & 81.16 & 67.21 \\
 \tStagedTraining{}-\mComplexFeatureFusion{} & 78.11 & 77.51 & 79.78 \\
 
 \bottomrule

\end{tabular}
\caption{Additional metrics for \textbf{Visual Aid} component - F1 binary, Precision and Recall.}
\vspace{-10.0pt}
\label{tab:add_res_vis_aid}
\end{table}

\begin{table}[h!]
\centering
\small
\begin{tabular}{lrrr}

\toprule
 Model & F1-binary & Precision & Recall  \\
 \midrule

 Text (Manual) & 94.95 & 95.16 & 94.90 \\
 Text (ASR) & 73.98 & 79.11 & 70.10 \\
 Audio (Emotion) & 64.31 & 66.40 & 62.41 \\
 Video & 54.35 & 59.88 & 52.20 \\
 \midrule
    \multicolumn{4}{c}{\textbf{Fusion using Manual Transcription}} \\
 \midrule

 \mVotingFusion{} & 74.87 & 79.72 & 70.70 \\[5pt]
 
 \tJointTraining{}-\mEnsembleFusion{} & 94.56 & 95.66 & 93.62 \\
 \tStagedTraining{}-\mEnsembleFusion{} & 96.18 & 96.94 & 95.54 \\[5pt]
 
 \tJointTraining{}-\mComplexFeatureFusion{} & 94.90 & 95.68 & 94.26 \\
 \tStagedTraining{}-\mComplexFeatureFusion{} & 96.16 & 96.89 & 95.54 \\

 \midrule
    \multicolumn{4}{c}{\textbf{Fusion using ASR Transcription}} \\
 \midrule

 \mVotingFusion{} & 71.09 & 69.27 & 73.24 \\[5pt]
 
 \tJointTraining{}-\mEnsembleFusion{} & 75.25 & 73.82 & 78.41 \\
 \tStagedTraining{}-\mEnsembleFusion{} & 75.42 & 79.29 & 72.04 \\[5pt]
 
 \tJointTraining{}-\mComplexFeatureFusion{} & 77.49 & 71.51 & 84.68 \\
 \tStagedTraining{}-\mComplexFeatureFusion{} & 74.27 & 80.01 & 69.45 \\
 
 \bottomrule

\end{tabular}
\caption{Additional metrics for \textbf{Praise} component - F1 binary, Precision and Recall.}
\vspace{-10.0pt}
\label{tab:add_res_praise}
\end{table}

\begin{table}[h!]
\centering
\small
\begin{tabular}{lrrr}

\toprule
 Model & F1-binary & Precision & Recall  \\
 \midrule

 Text (Manual) & 80.94 & 84.03 & 78.33 \\
 Text (ASR) & 67.88 & 73.00 & 63.67 \\
 Audio (Emotion) & 68.88 & 65.95 & 73.00 \\
 Video & 66.19 & 65.61 & 67.00 \\
 \midrule
    \multicolumn{4}{c}{\textbf{Fusion using Manual Transcription}} \\
 \midrule

 \mVotingFusion{} & 79.75 & 79.51 & 80.00 \\[5pt]
 
 \tJointTraining{}-\mEnsembleFusion{} & 81.42 & 82.62 & 80.33 \\
 \tStagedTraining{}-\mEnsembleFusion{} & 85.64 & 87.49 & 84.33 \\[5pt]
 
 \tJointTraining{}-\mComplexFeatureFusion{} & 81.60 & 82.60 & 80.67 \\
 \tStagedTraining{}-\mComplexFeatureFusion{} & 85.68 & 87.53 & 84.00 \\

 \midrule
    \multicolumn{4}{c}{\textbf{Fusion using ASR Transcription}} \\
 \midrule

 \mVotingFusion{} & 69.92 & 68.25 & 71.67 \\[5pt]
 
 \tJointTraining{}-\mEnsembleFusion{} & 70.03 & 71.32 & 69.00 \\
 \tStagedTraining{}-\mEnsembleFusion{} & 71.65 & 71.72 & 71.63 \\[5pt]
 
 \tJointTraining{}-\mComplexFeatureFusion{} & 67.58 & 69.98 & 66.00 \\
 \tStagedTraining{}-\mComplexFeatureFusion{} & 70.24 & 71.54 & 69.33 \\
 
 \bottomrule

\end{tabular}
\caption{Additional metrics for \textbf{Anatomic} component - F1 binary, Precision and Recall.}
\vspace{-10.0pt}
\label{tab:add_res_anatomic}
\end{table}

\begin{table}[h!]
\centering
\small
\begin{tabular}{lrrr}

\toprule
 Model & F1-binary & Precision & Recall  \\
 \midrule

 Text (Manual) & 70.39 & 68.02 & 73.33 \\
 Text (ASR) & 65.71 & 65.76 & 65.67 \\
 Audio (Emotion) & 64.89 & 60.29 & 70.67 \\
 Video & 63.88 & 64.10 & 63.67 \\
 \midrule
    \multicolumn{4}{c}{\textbf{Fusion using Manual Transcription}} \\
 \midrule

 \mVotingFusion{} & 72.04 & 71.83 & 72.33 \\[5pt]
 
 \tJointTraining{}-\mEnsembleFusion{} & 72.80 & 71.90 & 74.33 \\
 \tStagedTraining{}-\mEnsembleFusion{} & 77.16 & 75.19 & 79.67 \\[5pt]
 
 \tJointTraining{}-\mComplexFeatureFusion{} & 73.43 & 70.06 & 77.33 \\
 \tStagedTraining{}-\mComplexFeatureFusion{} & 76.75 & 75.66 & 78.33 \\

 \midrule
    \multicolumn{4}{c}{\textbf{Fusion using ASR Transcription}} \\
 \midrule

 \mVotingFusion{} & 64.56 & 63.23 & 66.00 \\[5pt]
 
 \tJointTraining{}-\mEnsembleFusion{} & 64.69 & 67.29 & 63.00 \\
 \tStagedTraining{}-\mEnsembleFusion{} & 72.89 & 69.59 & 76.67 \\[5pt]
 
 \tJointTraining{}-\mComplexFeatureFusion{} & 64.77 & 68.39 & 62.67 \\
 \tStagedTraining{}-\mComplexFeatureFusion{} & 67.39 & 66.00 & 69.33 \\
 
 \bottomrule

\end{tabular}
\caption{Additional metrics for \textbf{Procedural} component - F1 binary, Precision and Recall.}
\vspace{-10.0pt}
\label{tab:add_res_procedural}
\end{table}

\begin{table}[h!]
\centering
\small
\begin{tabular}{lrrr}

\toprule
 Model & F1-binary & Precision & Recall  \\
 \midrule

 Text (Manual) & 74.48 & 74.26 & 75.00 \\
 Text (ASR) & 63.98 & 66.86 & 65.00 \\
 Audio (Emotion) & 68.11 & 67.97 & 69.00 \\
 Video & 64.99 & 67.13 & 63.67 \\
 \midrule
    \multicolumn{4}{c}{\textbf{Fusion using Manual Transcription}} \\
 \midrule

 \mVotingFusion{} & 73.84 & 75.15 & 72.67 \\[5pt]
 
 \tJointTraining{}-\mEnsembleFusion{} & 74.85 & 74.18 & 75.67 \\
 \tStagedTraining{}-\mEnsembleFusion{} & 80.06 & 76.18 & 84.67 \\[5pt]
 
 \tJointTraining{}-\mComplexFeatureFusion{} & 74.28 & 80.88 & 69.00 \\
 \tStagedTraining{}-\mComplexFeatureFusion{} & 80.67 & 79.77 & 81.67 \\

 \midrule
    \multicolumn{4}{c}{\textbf{Fusion using ASR Transcription}} \\
 \midrule

 \mVotingFusion{} & 64.64 & 71.01 & 63.00 \\[5pt]
 
 \tJointTraining{}-\mEnsembleFusion{} & 65.48 & 69.70 & 67.00 \\
 \tStagedTraining{}-\mEnsembleFusion{} & 70.34 & 69.01 & 70.00 \\[5pt]
 
 \tJointTraining{}-\mComplexFeatureFusion{} & 65.15 & 66.52 & 67.67 \\
 \tStagedTraining{}-\mComplexFeatureFusion{} & 68.50 & 75.20 & 68.00 \\
 
 \bottomrule

\end{tabular}
\caption{Additional metrics for \textbf{Technical} component - F1 binary, Precision and Recall.}
\vspace{-10.0pt}
\label{tab:add_res_technical}
\end{table}

\end{document}